\documentclass[letterpaper, 10 pt, conference]{ieeeconf}  

\IEEEoverridecommandlockouts                              

\usepackage{amsmath,amsfonts,bm}

\def\eqref#1{equation~\ref{#1}}

\def\1{\bm{1}}

\def\va{{\bm{a}}}

\def\vs{{\bm{s}}}

\def\vx{{\bm{x}}}

\DeclareMathAlphabet{\mathsfit}{\encodingdefault}{\sfdefault}{m}{sl}
\SetMathAlphabet{\mathsfit}{bold}{\encodingdefault}{\sfdefault}{bx}{n}

\input{acronyms}

\usepackage{graphicx} 
\usepackage{xcolor}
\usepackage{tikzducks}
\usepackage[colorlinks=true,urlcolor=blue,linkcolor=black,citecolor=black]{hyperref} 
\usepackage[capitalize]{cleveref}
\usepackage{url}

\newtheorem{requirement}{Requirement}

\title{\LARGE \bf
Towards Safe Robot Foundation Models
}

\author{Maximilian T\"olle$^{\star, 1, 2}$\; Theo Gruner$^{\star, 1, 3}$\; Daniel Palenicek$^{\star, 1, 3}$\; Jonas G\"unster$^{\star, 1}$ \\ Puze Liu$^{1, 2}$\; Joe Watson$^{4}$\; Davide Tateo$^{1}$\; Jan Peters$^{1, 2, 3, 5, 6}$
\thanks{This work was supported by ``Third Wave of AI'', funded by the Excellence Program of the Hessian Ministry of Higher Education, Science, Research and Art. We also acknowledge the grant ``Einrichtung eines Labors des Deutschen Forschungszentrum für Künstliche Intelligenz (\textsc{dfki}) an der Technischen Universität Darmstadt'' of the Hessian Ministry of Science and Research, Arts and Culture.}
\thanks{*Equal contribution\; $^{1}$Technical University of Darmstadt\; $^{2}$German Research Center for Artificial Intelligence (\textsc{dfki})\; $^{3}$hessian.AI\; $^{4}$University of Oxford\; $^{5}$Robotics Institute Germany (\textsc{rig})\; $^{6}$Centre for Cognitive Science}%
\thanks{Correspondence: \newline \texttt{\{maximilian,theo,palenicek\}@robot-learning.de}}
}

\begin{document}

\maketitle
\thispagestyle{empty}
\pagestyle{empty}

\begin{abstract}

Robot foundation models hold the potential for deployment across diverse environments, from industrial applications to household tasks.
While current research focuses primarily on the policies' generalization capabilities across a variety of tasks, it fails to address safety, a critical requirement for deployment on real-world systems.
In this paper, we introduce a safety layer designed to constrain the action space of any generalist policy appropriately.
Our approach uses \acs{atacom}, a safe reinforcement learning algorithm that creates a safe action space and, therefore, ensures safe state transitions. By extending \acs{atacom} to generalist policies, our method facilitates their deployment in safety-critical scenarios without requiring any specific safety fine-tuning.
We demonstrate the effectiveness of this safety layer in an air hockey environment, where it prevents a puck-hitting agent from colliding with its surroundings, a failure observed in generalist policies.
\url{https://sites.google.com/robot-learning.de/towards-safe-rfm}
\end{abstract}


\section{Introduction}

\noindent Deploying autonomous agents in real-world environments requires motion generation that is both feasible and adaptable to various scenarios. \acp{rfm} advance this goal by being trained across a variety of embodiments, tasks, and environments. However, a critical component—safety—remains unaddressed despite its importance in many real-world applications.
Current \acp{rfm}~\cite{octo_2024, kimOpenVLAOpenSourceVisionLanguageAction2024} are typically trained with \ac{bc} to imitate expert trajectories. Given that expert data predominantly consists of safe demonstrations, \acp{rfm} may implicitly reflect a notion of safety as a result of this data bias. However, while this may encourage conservative behavior in safety-critical tasks, it does not provide any formal safety guarantees. Additionally, \ac{bc} policies may catastrophically damage the robot during deployment when encountering unseen observations due to the distribution shift \cite{invitation_to_imitation, osa2018algorithmic, dagger}. Therefore, we argue that integrating domain expertise is essential for ensuring reliable safety.

Inductive biases combat several shortcomings of purely data-driven approaches in robot learning~\cite{liu2022_atacom, diffusor_actor, funk2024actionflow, deep_lagrangian_networks, neural_posterior_domain_randomization}.
Incorporating analytic models into these optimization techniques enables sound inference~\cite{neural_posterior_domain_randomization}, data-efficient learning~\cite{deep_lagrangian_networks}, and safety~\cite{liu2022_atacom} by exploiting the problem's inherent dynamic structure. Safety has thereby been a major concern within the control community which has developed several safety-ensuring methods using \acl{cbf}~\cite{ames2019control, taylor2020learning, xiao2022high_order, tan2023your, yang2023model}, reachability analysis~\cite{akametalu2014reachability, fisac2018general, shao2021reachability, selim2022safe, wabersich2023data} and shielding~\cite{alshiekh2018safe, dalal2018safe, hewing2020learning, carr2023safe}.
Commonly, all these approaches exploit domain knowledge to construct a guaranteed safety filter.
For more details, refer to the following reviews ~\cite{safe_learning_in_robotics, data_driven_safety_filters, safety_filter_unified_view}.

In this paper, we introduce a safety module that enables a pre-trained \ac{rfm} to operate safely within an environment by adhering to domain-specific safety constraints. This is accomplished by utilizing system dynamics to control actions within the robot's null space.
Following \cite{liu2022_atacom, liu2024_atacom_journal}, we create a safe action space from the constraints and system dynamics, which ensures that an initially unsafe action from a \ac{rfm} is mapped to a safe action. By doing this, the module ensures safe transitions during deployment.

\begin{figure}[t]
    \centering
    \label{fig:octocom_scheme}
    \includegraphics[width=\linewidth]{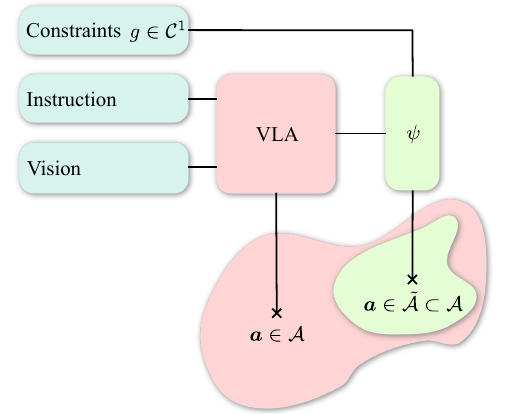}
    \caption{Composition of a safe \ac{vla} policy. The proposed safety module $\psi$ depends on a pre-specified set of safety constraints $g$ and can be added on top of the output of the \acs{vla} policy. While the initial output of the \ac{vla} policy does not guarantee safe actions, the added safety module ensures safe state transitions.}
    \vspace{-1.5em}
\end{figure}

\section{A Safety Module For Generalist Policies}

\noindent Today's \acp{rfm}~\cite{rt2_2023, open_x_embodiment_rt_x_2023, octo_2024, kimOpenVLAOpenSourceVisionLanguageAction2024} are trained on large-scale datasets \cite{open_x_embodiment_rt_x_2023} to predict actionable outputs from multi-modal observations $\vx$, such as images, language instructions and proprioceptive data. We define the policy that maps language instruction and observations to robotic actions as $\va \sim \pi_{\mathrm{\textsc{vla}}}(\cdot\mid\vx)$. 
Our goal is to make an already trained \ac{rfm} safe at test time. 
Therefore, we adopt \ac{atacom}~\cite{liu2024_atacom_journal} to ensure guaranteed safe actions.
We pose the following requirements for the system:
\begin{requirement}
    \label{req:affine_system}
    Access to the system's state $\vs$ and a control affine system $\dot\vs = f(\vs) + G(\vs)\va$.
\end{requirement}
\begin{requirement}
    \label{req:constraints}
    We define the safety conditions as continuously differentiable constraints $\bm{0} \geq g(\vx)\in\mathcal{C}^1$.
\end{requirement}

Most robotic manipulators in~\cite{open_x_embodiment_rt_x_2023} fulfill rigid-body assumptions and thus already comply with \cref{req:affine_system}. Furthermore, we assume that practitioners have prior knowledge of the robot’s safety requirements and can effectively define the system’s constraints (\cref{req:constraints}). Thus, while the above-stated requirements may seem restrictive at first, we deem that these assumptions hold for most currently considered robotic platforms \acp{vla} are trained on.

\vspace{0.5em}
\noindent\textbf{Acting on the tangent space of the constraint manifold.\quad}
Building on the aforementioned requirements, \ac{atacom} ~\cite{liu2022_atacom, liu2024_atacom_journal} constructs a constraint manifold of safe configurations.
Actions are then mapped into the tangent space of this manifold, ensuring safe transitions.
As such, \ac{atacom} can be seen as a mapping from actions $\va$, the state $\vs$, and the safety constraints $g$ to safe actions 
\begin{align*}
    \va_{\mathrm{safe}} = \psi_{G, f}(\va, \vs, g), \quad\va\sim\pi_{\mathrm{\textsc{vla}}}(\cdot\mid\vx).
\end{align*}
In this way, we ensure that actions that are drawn from a \ac{vla} policy are mapped to be safe actions that guarantee compliance with the safety constraints $g$.
%
%
\section{Experiments}
\noindent We empirically evaluate the proposed approach on a robot air hockey task. The objective is to hit a puck into the goal while adhering to multiple safety constraints, such as keeping the end-effector on the table surface, preventing the arm from colliding with the table, and ensuring joint position limits.  
We refer to \cite{liu2024_atacom_journal} for a detailed description of the experimental setup.
The policy's observation consists of language instructions, a goal image of the scene, and proprioceptive data in the form of joint positions, joint velocities, puck position, and puck velocity.
While not needed for safety but for improved performance in the air hockey task, we fine-tune a pre-trained \ac{octo} \cite{octo_2024} policy using behavior cloning in both a simulated \ac{mujoco}~\cite{todorov2012mujoco} environment and a real-world setting. Importantly, we obtain the fine-tuning data by an expert policy that does not leverage \ac{atacom}.      
The policy outputs desired end-effector velocities in the x-y plane of the table surface, which are converted to joint velocities using inverse kinematics. The \ac{atacom} layer then maps these joint velocities to safe ones before passing them to a joint-space controller. We compare our safety-aware approach to an unsafe baseline, where the joint-space controller directly executes the unfiltered joint velocities.

\begin{figure}[t]
    \centering
    \includegraphics[width=\linewidth]{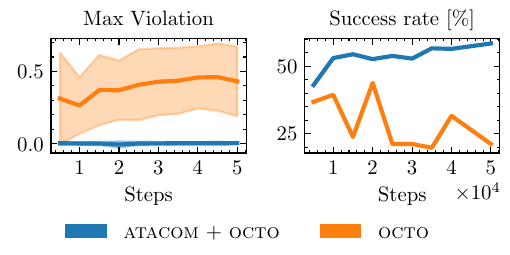}
    \caption{Safety violations of the \ac{octo} policy w/o the safety module on the air hockey hitting task for different checkpoints during the training phase. We report the maximum constraint violation a trajectory as well as the success rate of the robot hitting the puck into the goal. When the \ac{atacom} safety module is added, the policy remains compliant with safety constraints throughout fine-tuning. It progressively improves its success rate, whereas the unmodified \ac{octo} policy continues to breach safety limits while achieving a lower success rate.}
    \label{fig:results}
    \vspace{-1.5em}
\end{figure}

We evaluate the safety module for various fine-tuning checkpoints of \ac{octo} on the physical system. Several deployment videos of \ac{octo} playing air hockey can be found on our project page.
\cref{fig:results} shows that the \ac{octo} agent with the added safety module does not violate the safety constraints during inference. On the contrary, \ac{octo} without the added safety layer heavily violates the constraints even though the fine-tuning data contains safe expert demonstrations. Looking at the success rates, we observe a steady performance improvement in the number of training iterations when using the safety module. Importantly, while the fine-tuning data is not obtained with \ac{atacom}, we still obtain high success rates, which underlines that \ac{atacom} does not generate overly conservative control actions. Interestingly, we see that the constraint violations of the \ac{octo} baseline increase with added training time, which negatively impacts the policy performance. 

\section{Conclusion}
\noindent We propose a safety module that can be added as the final layer of a \acf{rfm} by leveraging domain-specific knowledge. Although leveraging domain knowledge may seem counterintuitive for \acp{rfm}, we hypothesize that reasoning with system dynamics is essential for ensuring safety. Additionally, by designing this module as an independent safety layer, it does not incorporate any additional computational burden, such as fine-tuning, to ensure safety. 
We demonstrate the effectiveness of the safety layer by evaluating a \ac{vla} policy with \ac{bc} on an air hockey hitting task for which it is critical not to crash with the tabletop. 
While we emphasize that ensuring safety requires domain expertise, it can also be a demanding task to formulate all scene-specific safety constraints. One intuitive research direction is to automate the process by leveraging the inherent knowledge of \acp{vlm}. However, so far, \acp{vlm} have only been used to integrate semantic safety constraints such as ``keep the cup upright" into an already existing set of constraints~\cite{updatingrobotsafetyrepresentations2024, semanticallysaferobotmanipulation2024}. Beyond the formulation of safety constraints, it remains an open research question of how a more generalizable concept of safety can be formulated and applied across different embodiments, environments, and tasks.    








\newpage

\addtolength{\textheight}{-9cm}   

\bibliographystyle{IEEEtran}
\bibliography{bibliography}

\end{document}